\def\year{2022}\relax
\documentclass[letterpaper]{article} 
\usepackage{aaai22}  
\usepackage{times}  
\usepackage{helvet}  
\usepackage{courier}  
\usepackage[hyphens]{url}  
\usepackage{graphicx} 
\usepackage{natbib}  
\usepackage{caption} 
\usepackage{amsmath,bm,times,amssymb}
\usepackage{soul}
\usepackage{xspace}

\DeclareCaptionStyle{ruled}{labelfont=normalfont,labelsep=colon,strut=off} 
\usepackage{algorithm}
\usepackage{algorithmic}
\usepackage{multirow}
\usepackage{algorithmic}

\usepackage{tikz}
\usepackage{pgfplots}
\pgfplotsset{width=7.5cm,compat=1.12}
\usepgfplotslibrary{fillbetween}
\usepackage{subcaption}
\usepackage{mathtools, nccmath}

\usepackage{newfloat}
\usepackage{listings}

\renewcommand{\th}[0]{\textsuperscript{th}\xspace}

\floatstyle{ruled}
\newfloat{listing}{tb}{lst}{}
\floatname{listing}{Listing}
\title{{Wasserstein Adversarial Transformer for Cloud Workload Prediction}}
\author{
    Shivani Arbat\textsuperscript{\rm 1}, Vinodh Kumaran Jayakumar\textsuperscript{\rm 2}, Jaewoo Lee\textsuperscript{\rm 1}, Wei Wang\textsuperscript{\rm 2}, In Kee Kim\textsuperscript{\rm 1} 
}
\affiliations{
    \textsuperscript{\rm 1}University of Georgia\\
    \textsuperscript{\rm 2} The University of Texas at San Antonio\\
    sga64681@uga.edu, rvn028@my.utsa.edu, jaewoo.lee@uga.edu,  wei.wang@utsa.edu, inkee.kim@uga.edu

}

\usepackage{bibentry}

\begin{document}

\maketitle

\begin{abstract}

Predictive {Virtual Machine (VM)} auto-scaling is a promising technique to optimize cloud applications' operating costs and performance. 
Understanding the job arrival rate is crucial for accurately predicting future changes in cloud workloads and proactively provisioning and de-provisioning VMs for hosting the applications. 
However, developing a model that accurately predicts cloud workload changes is extremely challenging due to the dynamic nature of cloud workloads.
Long-Short-Term-Memory (LSTM) models have been developed for cloud workload prediction. 
Unfortunately, the state-of-the-art LSTM model leverages recurrences to predict, which naturally adds complexity and increases the inference overhead as input sequences grow longer. 
To develop a cloud workload prediction model with high accuracy and low inference overhead, 
this work presents a novel time-series forecasting model called WGAN-gp Transformer, inspired by the Transformer network and improved Wasserstein-GANs. The proposed method adopts a Transformer network as a {\em generator} and a multi-layer perceptron as a {\em critic}. The extensive evaluations with real-world workload traces show WGAN-gp Transformer achieves $5\times$ faster inference time with up to $5.1\%$ higher prediction accuracy {against the state-of-the-art approach}.
We also apply WGAN-gp Transformer to auto-scaling mechanisms on Google cloud platforms, and the WGAN-gp Transformer-based auto-scaling mechanism outperforms the LSTM-based mechanism by significantly reducing VM over-provisioning and under-provisioning rates.

\end{abstract}

\section{Introduction}
Resource provisioning and VM auto-scaling of cloud resources are essential operations to optimize cloud costs and the performance of cloud applications~\cite{shen2011cloudscale, MingAutoscaling:SC11, Autoscale:Eurosys2020}. 
Auto-scaling dynamically performs scale-out and scale-in operations as application workloads fluctuate. {\em e.g.,} change in user requests to the applications. 
The scale-out operation increases the number of VMs for running the cloud application as the workload increases so that the cloud application leverages enough amount of VMs and meets its performance goal. 
On the other hand, the scale-in operation automatically downsizes the number of existing VMs by terminating idle VMs when the workload decreases and helps the cloud application minimize the cloud cost. 

While auto-scaling offers benefits to cloud applications, unavoidable delays occur during the VM scaling processes, {\em e.g.,} VM startup delay due to the reactive nature, hence offering suboptimal cloud resource management~\cite{VMStartupDelay:CLOUD12, VMStartupDelay:CLOUD21, VMStartupDelay:arXiv21}.
To address the reactive nature of auto-scaling, predictive auto-scaling approaches have been deeply investigated~\cite{shen2011cloudscale, roy2011efficient, ARIMAWorkPred-TCLOUD15, EmpEvalWLPred:CLOUD16, kim2018cloudinsight, jayakumar2020self}. Predictive auto-scaling mechanisms commonly have two components: {\em workload predictor} and {\em auto-scaling module}. 
The role of workload predictor is to forecast the changes in user requests (or workloads) to cloud applications. 
An essential step in designing the workload predictor is to understand job arrival rates\footnote{We use the terms workloads, user requests, and job arrival rates interchangeably.}(JARs).

\begin{figure*}
    \centering\includegraphics[width=\linewidth]{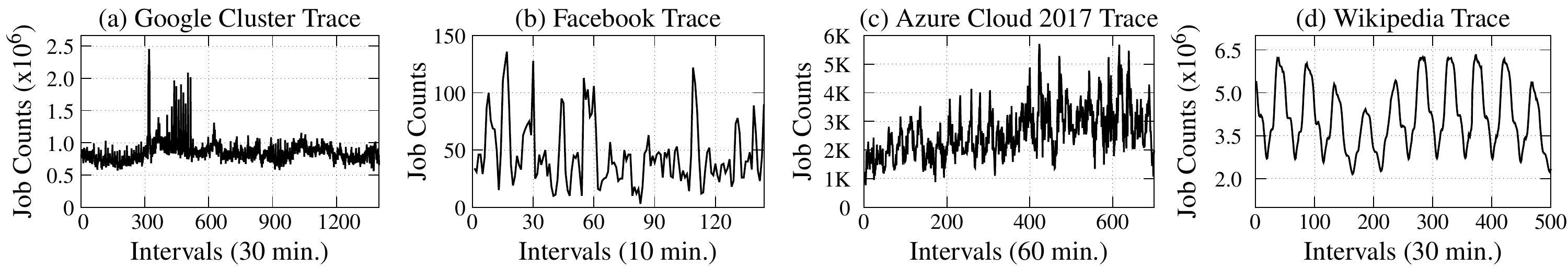}
    \vspace{-1em}
    \caption{Cloud Workload Traces. (a) Google Cluster Trace, (b) Facebook Hadoop Trace, (c) Azure Cloud 2017 VM Trace, and (d) Wikipedia Trace}
    \label{fig:cloud_workloads}
    \vspace{-1em}
\end{figure*}

There have been several approaches used in the workload predictor to model the JARs --- for example, statistical time-series methods~\cite{roy2011efficient, ARIMAWorkPred-TCLOUD15}, traditional machine learning~\cite{EmpEvalWLPred:CLOUD16, AzureResourceCentral-SOSP17}, ensemble learning~\cite{Vadara:cloudcom2014, ASAP:cloudcom2014, kim2018cloudinsight, CloudInsight:TCC20}, and deep learning~\cite{2019-Kumar-IJCNN, jayakumar2020self}.
The state-of-the-art approach in predicting cloud workloads employs a combination of LSTM and Bayesian optimization~\cite{jayakumar2020self}
which specifically leverage the power of LSTM to understand JARs using recurrences~\cite{1997-Hochreiter-LSTM}.
However, these recurrences increase complexity and computational overhead as input sequences grow longer.
Additionally, while LSTM is capable of detecting long-term seasonality in the cloud workloads, the majority of real-world cloud workloads have random and dynamic burstiness~\cite{WLBurstiness:SoCC15} as shown in cloud workload traces in Figure~\ref{fig:cloud_workloads}. 
The sudden spikes in cloud workload traces show unprecedented changes in user request patterns over time. 
Therefore, there is an urgent need to develop a novel cloud workload predictor to model such dynamic fluctuations and provide high prediction accuracy and low computational overhead.

In this work, we present a novel time-series forecasting model for cloud resource provisioning, called WGAN-gp (Wasserstein Generative Adversarial Network with gradient penalty) Transformer. WGAN-gp Transformer is inspired by Transformer networks~\cite{vaswani2017attention} and improved WGAN (Wasserstein Generative Adversarial Networks)~\cite{arjovsky2017wasserstein}. 
Our proposed model uses the transformer architecture for the {\em generator} and the {\em critic} is a multi-layer perceptron (MLP).
WGAN-gp Transformer also employs MADGRAD (Momentumized, Adaptive, Dual Averaged Gradient Method for Stochastic Optimization) as the model optimizer for both {\em generator} and {\em critic}
to achieve better convergence compared to widely adopted Adam optimizer~\cite{defazio2021adaptivity}. 
WGAN-gp Transformer is publicly available at \url{https://github.com/shivaniarbat/wgan-gp-transformer}.

We thoroughly evaluate the accuracy and overhead of WGAN-gp Transformer on the representative cloud workload datasets, and the evaluation results confirm that WGAN-gp Transformer consistently performs better, yielding lower prediction errors against the state-of-the-art LSTM model~\cite{jayakumar2020self}.
WGAN-gp Transformer achieves up to $5.1\%$ lower prediction error and $5\times$ faster prediction time against the baseline.

\section{Related Work}
\subsubsection{Statistical and Machine Learning (ML) Approaches}

Various methods have been applied to develop workload predictors in predictive VM auto-scaling. The statistical and ML models include Exponential Smoothing, Weighted Moving Average, Autoregressive models (AR) and variations ({\em e.g.,} ARMA, ARIMA), Support Vector Machine, Random Forest, Gradient Boosting, and others~\cite{shen2011cloudscale,wood2008profiling, roy2011efficient, ARIMAWorkPred-TCLOUD15, Wrangler-SoCC14, lin2015workload, AzureResourceCentral-SOSP17, Autoscale:Eurosys2020}.
While such statistical and ML approaches could model time-series cloud workloads with cyclic or seasonal trends, such approaches appeared to make sub-optimal predictions for cloud workloads as they were continuously and dynamically changing over time.
Moreover, a model can be effective on one ``{\em known}" type of workload, but it often fails to accurately predict future changes in other ``{\em previously unknown}" (previously not trained) workload patterns~\cite{EmpEvalWLPred:CLOUD16}.

\subsubsection{Neural Networks (NN)-based Approaches} 

The applications of NNs in time-series forecasting can provide improved accuracy across multiple domains. NN models learn to encode relevant historical information from time-series data into intermediate feature representation to make a final forecast with series of non-linear NN layers~\cite{lim2021time}.
For cloud workloads, LSTM~\cite{1997-Hochreiter-LSTM} and its variations are studied to forecast the resource demands or user requests.~\cite{2019-Nguyen-CCGrid, 2019-Kumar-IJCNN,jayakumar2020self}. 
In particular, Jayakumar et al.~\cite{jayakumar2020self} proposed LoadDynamics, a self-optimized generic workload prediction framework for cloud workloads. 
LoadDynamics performs autonomous self-optimization operations using Bayesian Optimization to repeatedly find the proper hyperparameters in LSTM for handling dynamic fluctuations of cloud workloads.
However, LSTM intrinsically depends on capturing long/short dependencies using recurrences. 
As the input sequence length grows, it increases the complexity of processing such longer input sequences. 

Due to the recent advancement of in the field of natural language processing, attention mechanism in Transformer network can be an alternative to recurrences or convolutions~\cite{vaswani2017attention}.
For example, TransGAN~\cite{jiang2021transgan} proposed a strict, Transformer-based GAN, which employs Transformer as both generator and discriminator. 
For time-series data, the attention mechanism in Transformer network allows the model to focus on temporal information in the input sequences~\cite{lim2021time}.
Moreover, Adversarial Sparse Transformer (AST)~\cite{wu2020adversarial} was proposed to leverage a sparse attention mechanism for increasing the prediction accuracy at the sequence level. This approach employed sparse transformer as generator and MLP as discriminator. 
Unfortunately, AST still has limitations to accurately predict highly dynamic cloud workloads because AST often loses long-term forecasting information due to the difficulty of training GANs with sparse point-wise connections.
    
On the other hand, to effectively predict dynamic cloud workloads with capturing long-term temporal information, our method proposes to train the Transformer network using an improved WGAN-gp algorithm~\cite{gulrajani2017improved}.

\section{Background}
    
\subsubsection{Problem definition}

A univariate time-series is defined as a sequence of measurements of the same variable collected over time. 
We study univariate time-series data of JARs or user requests rates collected at regular time intervals from various cloud workloads.  
Let $\mathbf{x} = [x_1, x_2, \ldots, x_T]$ denote  a univariate time-series of length $T$, where $x_t \in \mathbb{R}$ is its value at time $t$. We use boldface roman lower and upper case letters to denote vectors and matrices, respectively. $\mathbf{x}_{k:\ell}$ denotes the entries with indices from $k$ to $\ell$. To prepare a training dataset $\mathbf{X}\in \mathbb{R}^{N\times S}$, $\mathbf{x}_{1:T}$ is split into $N$ time series of length $S$. We write $\mathbf{x}_{i, 1:S}$ to denote the $i$\th time series in $\mathbf{X}$.

\subsubsection{Generative Adversarial Networks} \label{GAN-section}

Generative Adversarial Networks (GAN) \cite{goodfellow2014generative} are \textit{adversarial nets} framework, which simultaneously trains two completing models; a \textit{generative} model ($G$) and a \textit{discriminative} model ($D$). 
The training strategy leverages a min-max game between two competing models ($G$ and $D$), and the value function $V(D,G)$ is defined as 

\begin{equation}\label{eq:1} 
  \begin{split}
    \underset{G}{\min} \  \underset{D}{\max}\ V(D,G) =  \mathbb{E}_{x \sim \mathbb{P}_{r}} [\log D(x)]\  + \\   \mathbb{E}_{\tilde{x} \sim \mathbb{P}_{g}} [\log(1-D(\tilde{x}))],
  \end{split}
\end{equation}
where $\mathbb{P}_{r}$ is the real data distribution, $\mathbb{P}_{g}$ is the model distribution implicitly defined by $\tilde{x}$ = $G(z)$, and $z\sim p(z)$ is a latent variable having a simple distribution such as uniform distribution or standard normal distribution.
    
\subsubsection{Wasserstein GAN} 

Arjovsky et al.~\cite{arjovsky2017wasserstein} showed that the gradient of Jensen-Shannon divergence used in the original GAN is not smooth and may not be well defined when the model distribution and the true distribution have different supports. To mitigate the issue, Arjovsky et al. proposed Wasserstein GAN (WGAN), which consists of a generator and a critic.
In WGAN, the role of the generator is to generate a sample as in the original GAN, while that of the critic is to approximate the Wasserstein distance between $\mathbb{P}_{r}$ and $\mathbb{P}_{g}$. 
The objective function of WGAN is formulated by

\begin{equation} \label{eq:WGAN-val-func} 
    \underset{G}{\min} \  \underset{D \in \mathcal{D} }{\max}\ \mathbb{E}_{x \sim \mathbb{P}_{r}} [D(x)] - \mathbb{E}_{\tilde{x} \sim \mathbb{P}_{g}} [D(\tilde{x})],
\end{equation} 
where $\mathcal{D}$ is a set of 1-Lipschitz functions, $\mathbb{P}_{r}$ represents the data distribution, and $\mathbb{P}_{g}$ is the model distribution implicitly defined by $\tilde{x}$ = $G(z)$ with a latent variable $z \sim p(z)$.

In~\cite{arjovsky2017wasserstein}, the Lipschitz constraint on $\mathcal{D}$ is enforced by clipping the weight values into a small interval $[-c, c]$. However, it is unknown how to choose the hyperparameter $c$ that has a significant impact on the training of WGAN.
Furthermore, irregular value surfaces are generated due to hard clipping of the magnitude of each weight. 
Other weight constraints ({\em e.g.,} L2 norm clipping, weight normalization) and soft constraints ({\em e.g.,} L1 and L2 weight decay) also lead to similar problems~\cite{gulrajani2017improved}. 

\subsubsection{Gradient Penalty} 

To address the drawbacks of weight clipping, Gulrajani et al.~\cite{gulrajani2017improved} proposed an alternative way to enforce Lipschitz constraint. From the observation that functions $f$ with $\|f(x)\|\leq L$ are $L$-Lipschitz, they proposed to add a penalty term that forces the gradient norm of critic to stay close to 1, which results in the following objective function.

\begin{equation}\label{eq:gradient-penalty}
  \begin{split} 
    L = \mathbb{E}_{\widetilde{x} \sim \mathbb{P}_{g}} [D(\widetilde{x})] - \mathbb{E}_{x \sim \mathbb{P}_{r}} [D(x)]   + \\
    \lambda \mathbb{E}_{\overline{x} \sim \mathbb{P}_{\overline{x}}} [(|| \nabla_{\overline{x}}D(\overline{x})||_{2} - 1)^{2}]\,,
  \end{split}
\end{equation}
where $\overline{x} = \epsilon x + (1 - \epsilon)\widetilde{x} $ and $\epsilon \sim U[0,1].$
{The gradient penalty coefficient $\lambda$ controls how strictly the constraint is.} In our implementation, we set  
$\lambda = 10$ (the default values suggested by the authors of WGAN-GP).

\section{Wasserstein Adversarial Transformer}
\begin{figure*}[t]
    \centering\includegraphics[width=0.75\linewidth]{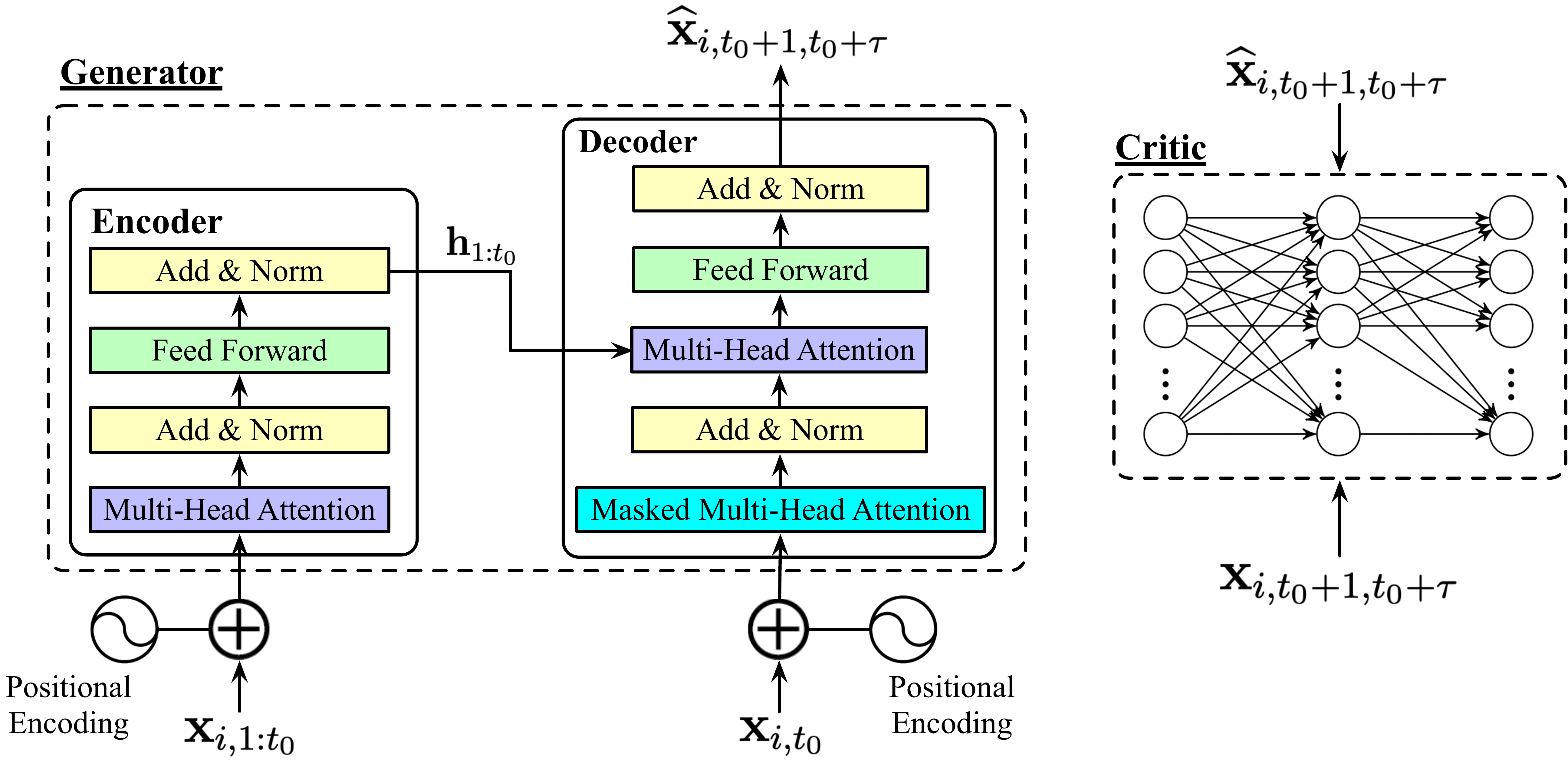}
    \caption{The architecture of WGAN-gp Transformer}
    \label{fig:wgan-generator}
\end{figure*}

As discussed in the Introduction, cloud workloads are often bursty and dynamically fluctuating over time. 
To accurately forecast future changes in cloud workloads, we propose {WGAN-gp Transformer}, based on the 
Transformer network and WGAN. 

Given an input time series $\mathbf{x}_{i, 1:S}$, $i=1, \ldots, N$, our model $G(\cdot\,; \bm{\theta})$ with parameter $\bm{\theta}$ predicts the values for time steps in $[t_0+1, t_0+\tau]$ conditioning on $\mathbf{x}_{i,1:t_0}$, where $t_0 + \tau = S$. $\tau$ is the number of time steps $G(\cdot\,; \bm{\theta})$ is trained to predict. That is, $\widehat{\mathbf{x}}_{t_0+1:t_0+\tau} = G(\mathbf{x}_{1:t_0}; \bm{\theta})$. The time ranges $[1, t_0]$ and $[t_0+1:S]$ are referred to as conditioning range and prediction range, respectively.

\subsubsection{Architecture of WGAN-gp Transformer}  
Figure~\ref{fig:wgan-generator} illustrates the architecture of the proposed WGAN-gp Transformer. 
The \textit{critic} network is composed of 3 fully connected layers with \texttt{LeakyReLU} as an activation function~\cite{xu2015empirical, wu2020adversarial}.
The {\em generator} is based on the encoder-decoder architecture of Transformer network, which is composed of one layer of an encoder and subsequent one layer of a decoder.
The encoder encodes the input time series $\mathbf{x}_{i, 1:t_0}$ into the latent vector $\mathbf{h}_{1:t_0}$.
The latent vector $\mathbf{h}_{1:t_0}$ serves as a memory for the decoder to generate values in the prediction range.
Transformer networks are order-agnostic~\cite{vaswani2017attention, tsai2019transformer} and, unlike recurrences and convolutions, it does not have implicit mechanisms to retain temporal (positional) dependency information.
To embed the temporal information, {positional encoding~\cite{vaswani2017attention}} is applied to the input $\mathbf{x}_{i, 1:t_0}$.
Similarly, positional encoding is applied to the input to decoder. 
The input to decoder is $\mathbf{x}_{i,t_0}$, which is the last time step value of $\mathbf{x}_{i, 1:t_0}$.

\subsubsection{Adversarial Training}

The generator (the Transformer network) attempts to generate high quality predictions that look similar to original time series by minimizing the mean absolute error $\mathcal{L}_{\text{MAE}}=\frac{1}{N}\sum_{i=1}^N \sum_{t=t_0+1}^{S} |\widehat{x}_{i,t} - x_{i,t}|$.
The objective function for generator is given by
\begin{align}
L_G 
&= \frac{1}{N}\sum_{i=1}^{N}     \|\widehat{\mathbf{x}}_{i,t_0+1:S} - \mathbf{x}_{i,t_0+1:S}\|_1  \nonumber \\
&\qquad -\frac{1}{N}\sum_{i=1}^{N} D([\mathbf{x}_{i, 1:t_0} \oplus \widehat{\mathbf{x}}_{i,t_0+1:S}])\,, \label{eq:wgan-gp-updated-generator-loss}
\end{align}
where $[\mathbf{a} \oplus \mathbf{b}]$ denotes the concatenation of two vectors.

\texttt{LeakyReLU} is computed by $f(x)=\max(\alpha x, x)$, where $\alpha=0.2$. 
The \textit{critic}'s objective function is given as follows.
\begin{align} \small
L_C
&= \frac{1}{N}\sum_{i=1}^{N} \Bigl\{
D([\mathbf{x}_{i, 1:t_0} \oplus \widehat{\mathbf{x}}_{i,t_0+1:S}])\Bigr\}    
- \frac{1}{N}\sum_{i=1}^{N} D(\mathbf{x}_{i,1:S}) \nonumber \\ 
& \qquad +  \lambda \frac{1}{N}\sum_{i=1}^{N} \left((\|\nabla_{\overline{x}_i}D(\overline{x}_i)\|_{2} - 1)^{2} \right)\,,
\label{eq:wgan-gp-updated-critic-loss}
\end{align}
where $\overline{x}_i = \epsilon\, \mathbf{x}_{i, 1:S} + (1-\epsilon)[\mathbf{x}_{i, 1:t_0} \oplus \widehat{\mathbf{x}}_{i,t_0+1:S}]$ and $\epsilon \sim U[0,1].$

\begin{algorithm}[t] \label{algo:wgan-gp-updated}
    \caption{Training Algorithm of WGAN-gp Transformer}
    \textbf{Require:} $\lambda$, the gradient penalty coefficient. $m$, the batch size. $n_{critic}$, the number of iterations of the critic per generator iteration.  $m_{MADGRAD}$, momentum value. $\alpha$ learning rate\\
    \textbf{Require:} initial critic parameters $w_{0}$, initial generator parameters $\theta_{0}$.
    \begin{algorithmic}[1]
    \WHILE{$\theta$ has not converged}
    \FOR{$t = 1,....,n_{critic}$}
    \FOR{$i = 1,....,m$}
    \STATE Sample \textit{real} data 
    \STATE Sample \textit{generator} output
    \STATE Compute $L_C$ 
    \ENDFOR
    \STATE $w \leftarrow$ \texttt{MADGRAD}($\nabla_{w}\frac{1}{m} \sum_{j=1}^{m}L_C,w,\alpha, m_{MADGRAD}$)
    \ENDFOR
    \STATE Sample a batch of generator output 
    \STATE $\theta \leftarrow$ \texttt{MADGRAD}($\nabla_{\theta}\frac{1}{m} \sum_{j=1}^{m} L_G, \theta, \alpha, m_{MADGRAD}$)
    \ENDWHILE
    \end{algorithmic}
\end{algorithm}
    
Algorithm 1 describes the training of WGAN-gp Transformer, and it is improved over the original proposal~\cite{gulrajani2017improved} to train the {\em generator}.
The algorithm updates 
the {\em critic} first and then updates the {\em generator} with learning results from the {\em critic}. 
The {\em generator} is trained with the updated loss function shown in Equation~\ref{eq:wgan-gp-updated-generator-loss} and the {\em critic} is trained with the updated loss function expressed in Equation~\ref{eq:wgan-gp-updated-critic-loss}.
We employ MADGRAD optimizer to minimize the updated loss function with gradient penalty for both {\em generator} and {\em critic}. 
MADGRAD is based upon the dual averaging formulation of AdaGrad for the model optimization~\cite{defazio2021adaptivity}.

\section{Experiments Setup}
\begin{table}[t] \small
\centering
\resizebox{1\columnwidth}!{
\begin{tabular}{l|c|c}
\hline
\textbf{Workload} & \textbf{Dataset Type} & \textbf{Time Interval (in $mins$)} \\ \hline \hline
Facebook & \multirow{3}{*}{Data Center} & \multirow{2}{*}{5, 10} \\ \cline{1-1}
Alibaba-2018 &  &  \\ \cline{1-1} \cline{3-3} 
Google &  & 10, 30 \\ \hline
Wiki & Web & 10, 30 \\ \hline
Azure-VM-2017 & \multirow{3}{*}{Cloud} & \multirow{2}{*}{10, 30, 60} \\ \cline{1-1}
Azure-VM-2019 &  &  \\ \cline{1-1} \cline{3-3} 
Azure-Func-2019 &  & 30, 60 \\ \hline
\end{tabular}
}
\caption{Cloud workload datasets}
\label{table:workload-time-interval}
\vspace{-1em}
\end{table}

\subsubsection{Workload Datasets}
Seven cloud workloads collected from different application categories are used to evaluate WGAN-gp Transformer. 
The workload traces are described in Table~\ref{table:workload-time-interval}. 
Facebook~\cite{chen2011case}, Alibaba (2018)\footnote{\url{https://github.com/alibaba/clusterdata}}, and Google~\cite{clusterdata:Wilkes2011} traces are from data center workloads.
Wikipedia workloads are from Wikibench\footnote{\url{http://www.wikibench.eu/}}. 
Three Azure workloads\footnote{\url{https://github.com/Azure/AzurePublicDataset/}} (Azure-VM-2017, Azure-VM-2019, Azure-Func-2019) are from cloud VM and function (serverless) services. 
These seven workloads are chosen because they have unique characteristics and dynamics in the workload patterns. 
For example, the data center workloads like Facebook and Google show dynamic spikes and high fluctuations in JARs.
Wikipedia dataset represents the behavior of web applications and strong seasonal patterns. 
Cloud VM and function workloads from Azure show a mixture of characteristics having high fluctuations and seasonality. 
Among seven selected workloads, Facebook, Google, Azure-VM-2017, and Wikipedia workloads are shown in Figure~\ref{fig:cloud_workloads}.
We omit figures of the other three workload patterns due to the page limitation. 

Different time granularities can exhibit subtle variations in the time-series workload patterns. 
Thus, with seven selected workloads, we generate 15 different {\em workload configurations} with different time intervals described in Table~\ref{table:workload-time-interval}. 
We use 5 and 10 minutes of the time interval for Facebook and Alibaba workloads. 10 and 30 minutes of the time interval are used for Google and Wikipedia workloads. 10, 30, and 60 minutes of the time interval are used for two Azure-VM workloads. Azure-Func-2019 workload uses two-time interval configurations with 30 and 60 minutes.

\begin{table}[t]\small
\centering
\resizebox{1\columnwidth}!{
\begin{tabular}{l|c|c|c|c}
\hline
\textbf{Workload} & \textbf{\begin{tabular}[c]{@{}c@{}}History\\ Len. ($n$)\end{tabular}} & \textbf{\begin{tabular}[c]{@{}c@{}}Batch \\Size ($m$)\end{tabular}} & \textbf{$d_{model}$} & \textbf{$n_{head}$} \\ \hline \hline
Facebook & {[}3-46{]} & {[}16-256{]} & \multirow{7}{*}{\begin{tabular}[c]{@{}c@{}}{[}8, 16, 32, 64, \\ 128, 512{]}\end{tabular}} & \multirow{7}{*}{[4, 8]} \\ \cline{1-3} 
Alibaba-2018 & {[}20-324{]} & \multirow{5}{*}{{[}16-1024{]}} &  &  \\ \cline{1-2} 
Google & {[}28-676{]} &  &  &  \\ \cline{1-2} 
Wikipedia (Wiki) & {[}12-274{]} &  &  &  \\ \cline{1-2} 
Azure-VM-2017 & {[}14-682{]} &  &  &  \\ \cline{1-2} 
Azure-VM\_2019 & {[}14-230{]} &  &  &  \\ \cline{1-3} 
Azure-Func-2019 & {[}7-108{]} & {[}16-512{]} &  &  \\ \hline
\end{tabular}
}
\caption{Hyperparameter search space for \textit{generator} in WGAN-gp Transformer}
\label{table:hyperparameter-search-space-WGAN-gp-Transformer}
\vspace{-1em}
\end{table}     

\subsubsection{Implementation}

WGAN-gp Transformer is implemented by using \texttt{PyTorch} and \texttt{scikit-learn}.
When training the generator and critic, we use the following configurations; $n_{critic} = 5$, $\lambda = 10$, and $\alpha$ (learning rate) $= 0.001$. 
We set the length of prediction range to $\tau = 1$ ({\em i.e.,} one-step ahead forecasting).
The MADGRAD optimizer employs the following default configurations; $m$ (momentum value) $=0.9$, $weight\_decay=0$, and {$eps=1e-6$}. 
And the training is performed with $1000$ epochs, which works well for our proposed method. 

When training and testing WGAN-gp Transformer, we divide the workload dataset sequence into three sub-datasets, containing the first $60\%$, the next $20\%$, and the remaining $20\%$, which are used for the model training, cross-validation, and testing, respectively.
We apply a sliding window approach to prepare the input data to WGAN-gp Transformer to divide the data into sequences of (history) length $n$. The model is trained to predict JAR at the next time step; thus, the sliding window moves with a stride of one-time step to acquire the input sequences. 
Finally, WGAN-gp Transformer is trained and evaluated on {NVIDIA GeForce RTX 2080 Ti} GPU machines.

\subsubsection{Hyperparameters}
We use an effective grid search to find the optimal parameters for training WGAP-gp Transformer. 
We use the configurations described in Table~\ref{table:hyperparameter-search-space-WGAN-gp-Transformer} for the hyperparameter search. 
$n$ (history length) is the length of the input sequence to the model,
$m$ is batch size,
$d_{model}$ is the number of input features for the encoder and decoder, and $n_{head}$: the number of heads in multi-head attention layer in encoder and decoder.
Please note that the model size ($d_{model}$) used for training the generator is the same number of input features in the critic linear layers.

\subsubsection{Evaluation metric}
We use Mean Absolute Percentage Error (MAPE) as the accuracy metric to assess the proposed method against the baseline. MAPE is calculated as,
$100 \times \left(\frac{1}{n}\right)\sum_{i=1}^{n}\left | \frac{\tilde{y}_{i} - y_{i}}{y_{i}} \right |$,
where $n$ is the total number of data points, $\tilde{y}_{i}$ represents predicted JAR at time step $i$, and $y_{i}$ represents actual JAR at time step $i$.

\subsubsection{Baseline}

WGAN-gp Transformer is evaluated against a state-of-the-art LSTM model, called LoadDynamics~\cite{jayakumar2020self}.
LoadDynamics employs LSTM model to automatically optimize LSTM for individual workload using Bayesian Optimization. As our baseline, we use the brute force version of LoadDynamics, which performs hyperparamter search for LSTM in predefined hyperparameter search space. 
For LoadDynamics, we use the same configurations for the model hyperparameters (the number of LSTM layers, the memory cell $C$ size, and the input length $n$) described in the original version of the paper. 
For the new cloud workloads (not been evaluated in the original LoadDyanmics paper), we use grid search to find the optimal hyperparameters. 
The training and testing of LoadDynamics are performed on the same GPU (NVIDIA GeForce RTX 2080 Ti) machine.

\section{Evaluation Results}
\begin{table}[!t]
\centering
\resizebox{0.90\columnwidth}!{
\begin{tabular}{l | c | c }
\hline
\bf {Workload} & \begin{tabular}[c]{@{}c@{}}\bf{Load-}\\ \bf{Dynamics}\end{tabular} & \begin{tabular}[c]{@{}c@{}}\bf{WGAN-gp}\\ \bf{Transformer}\end{tabular}  \\

\hline
\hline
Facebook-5m & 47.20 & \bf{42.11}  \\
\hline
Facebook-10m & 43.68  & \bf{39.31} \\
\hline
Alibaba-2018-5m & 17.95 & \bf{15.76} \\
\hline
Alibaba-2018-10m & 16.90 & \bf{14.67} \\
\hline
Google-10m & 11.49 & \bf{10.58} \\
\hline
Google-30m & 9.12 & \bf{8.34} \\
\hline
Wiki-10m & \bf{1.17} & 1.34 \\
\hline
Wiki-30m & \bf{1.75} & 3.43 \\ 
\hline
Azure-VM-2017-10m & 42.63 & \bf{41.32} \\
\hline
Azure-VM-2017-30m & 29.35 & \bf{27.48} \\
\hline
Azure-VM-2017-60m & 16.11 & \bf{12.77} \\
\hline
Azure-VM-2019-30m & 19.74 & \bf{15.19} \\
\hline
Azure-VM-2019-60m & 13.5 & \bf{10.82} \\
\hline
Azure-Func-2019-5m & \bf{1.63} & 3.05 \\ 
\hline
Azure-Func-2019-10m & 2.06 & \bf{1.85} \\ 
\hline
\end{tabular}
}
\caption{Average prediction errors (MAPE) for cloud workloads} \label{table:evaluation-results}
\vspace{-1.0em}
\end{table}

\subsubsection{Prediction Errors and Inference Overheads}
We first evaluate the prediction error of WGAN-gp Transformer. 
Table~\ref{table:evaluation-results} reports the prediction error of WGAN-gp Transformer and the baseline (LoadDynamics) when predicting 15 workload configurations.
While the prediction errors vary with different workload configurations, the results clearly show that WGAN-gp Transformer outperforms the baseline model for most workload configurations.
In particular, WGAN-gp Transformer provides up to $5.1\%$ lower prediction errors (MAPE). 
WGAN-gp Transformer relies on the alternative adversarial training techniques to enforce Lipschitz constraint using gradient penalty with MADGRAD optimizer. 
Additionally, the alternative adversarial training technique is able to reduce the prediction error against the LSTM's recurrences. 

\begin{figure}[t]
    \centering\includegraphics[width=\linewidth]{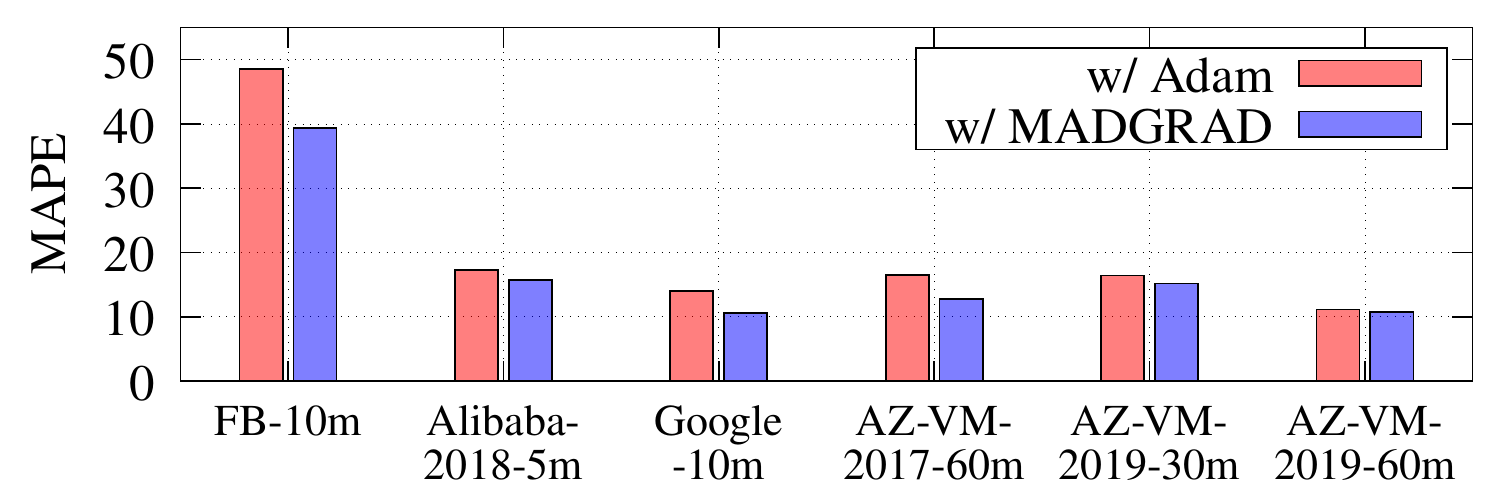}
    \caption{Prediction error comparison between WGAN-gp Transformer with Adam and WGAN-gp Transformer with MADGRAD optimizer. (FB: Facebook, Az: Azure)}
    \label{fig:diff_optimizer}
    \vspace{-1em}
\end{figure}

We examine the impact of the model optimizer on the accuracy of WGAN-gp Transformer by comparing the prediction errors of our model with both MADGRAD and Adam optimizer.
In this evaluation, Adam optimizer employs the following parameters;  $\beta_1 = 0$, $\beta_2 = 0.9$, and learning rate = $0.0001$ for both generator and critic. 
As shown in Figure~\ref{fig:diff_optimizer}, WGAN-gp Transformer with MADGRAD optimizer shows a significant reduction in the prediction errors, indicating that the use of MADGRAD optimizer is the critical factor for more accurate workload prediction. 

We also notice that WGAN-gp Transformer can be less accurate than LSTM-based forecasting for three workload datasets, {\em i.e.,} Wiki-10min, Azure-Func-2019-5m, and Azure-Func-2019-10m.
Our further analysis reveals that these workload patterns have relatively stronger seasonality than others. {\em e.g.,} Wikipedia workloads shown in Figure~\ref{fig:cloud_workloads}(d).
LSTM's memory capability can be better to store more accurate information for such repeating patterns and yield lower prediction errors. 

\begin{figure}[t]
    \centering\includegraphics[width=\linewidth]{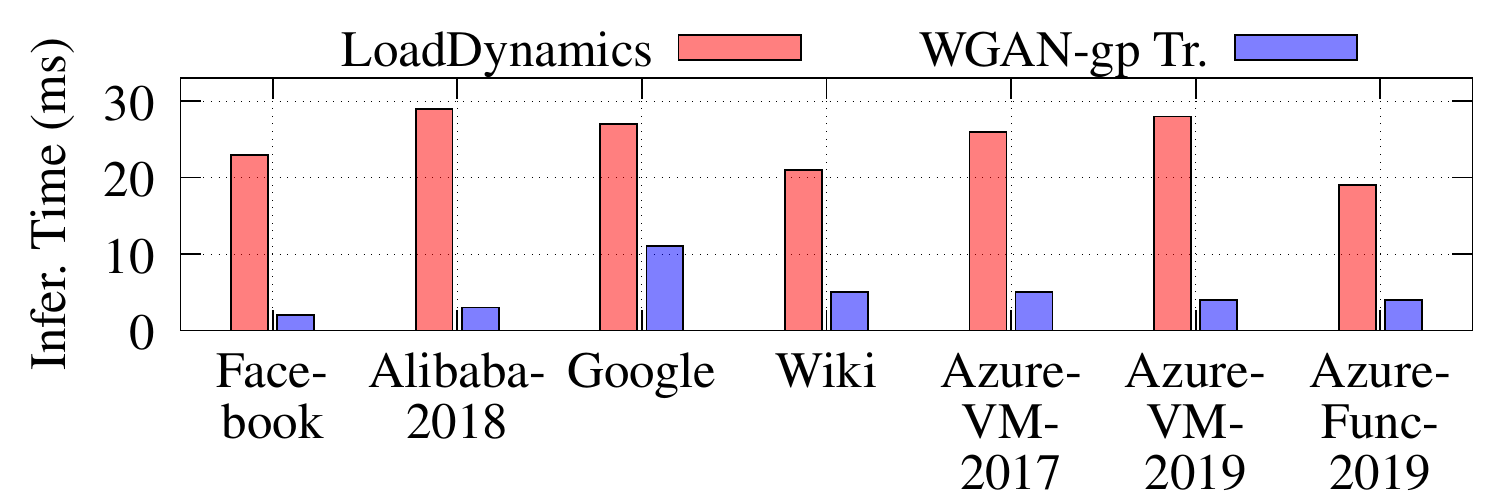}
    \caption{Inference time comparison. (Az: Azure)}
    \label{fig:inference-time-comparison}
    \vspace{-0.5em}
\end{figure}

We also measure the inference overhead (time) of both models when performing the prediction tasks. 
Figure~\ref{fig:inference-time-comparison} shows the inference time comparison between WGAN-gp Transformer and LSTM and represents the benefit of WGAN-gp Transformer over the LSTM-based model.
The results report that WGAN-gp Transformer has $5\times$ faster inference time over the baseline. 
On average, the average inference time of WGAN-gp is $4.85ms$, on the other hand, the average inference time of LoadDynamics is $25.57ms$. 
The faster inference time of WGAN-gp Transformer is because the model processes the input sequence at once, which results in quicker inference time. On the other hand, when LSTM processes the input in the sequence, it processes only one time step at a time, which increases the inference time for processing longer sequences in the prediction tasks.

\subsubsection{Auto-Scaling Evaluations on Google Cloud}

After evaluating the prediction errors and inference overhead, we develop VM auto-scaling systems with WGAN-gp Transformer and LoadDynamics and deploy them on a real-world cloud platform (GCP\footnote{\url{https://cloud.google.com/}}; Google Cloud Platform) to measure the performance differences in both systems. 

The auto-scaling systems consist of a {\em workload predictor} (WGAN-gp Transformer or LoadDynamics) and a {\em VM auto-scaler}. 
The workload predictor determines the predicted job arrivals and provides the prediction results to the auto-scaler. 
Suppose $P_{i}$, predicted at $(i-1)^{th}$ time interval, is the predicted number of jobs arriving at $i^{th}$ time interval to the auto-scaling system and represents the number of VMs to be created in advance. Note that we use an assumption that one job needs an allocation of a single VM.
Therefore, with $P_{i}$, the auto-scaler creates the $P$ VMs at $i^{th}$ time interval.
Assume $T_{i}$ is the actual number of jobs arriving at $i^{th}$ time interval. If $T_{i} > P_{i}$ (the actual job arrivals are greater than the predicted), then it results in {\em under-provisioning} and, to accommodate extra demand of the jobs, more VMs need to be allocated. 
In this case, additional time will be needed to finish the jobs due to the VM startup time~\cite{VMStartupDelay:CLOUD11}. 
On contrary, if $T_{i} < P_{i}$ (actual job arrivals are smaller than the predicted), this results in {\em over-provisioning} and incurs extra unnecessary cost with the VMs being idle. 
          
Google Cloud's \texttt{e2-medium} VMs are used for this evaluation. Facebook and Azure 2019 workloads are evaluated to compare the auto-scaling performance. 
For the evaluation with the Facebook workload, we use Cloud Suite's {\em Data Analytics} benchmark in CloudSuite, which performs large amounts of machine learning tasks using MapReduce framework~\cite{ferdman2012clearing}. 
For the evaluation with Azure 2019 VM workload, we use {\em In-Memory Analytics} benchmark in CloudSuite, which uses Apache Spark to execute collaborative filtering algorithm in-memory on dataset of user-movie ratings~\cite{ferdman2012clearing}. 
The evaluation results are reported in Table~\ref{table:resource-provisioning-facebook}. 
As shown in the results, the auto-scaling system with WGAN-gp Transformer outperforms the auto-scaling system with the baseline model.
With the Facebook workload, the auto-scaler with WGAN-gp Transformer shows a significant reduction in under-provisioning by $27.95\%$.
For the Azure VM 2019 workload, the system with WGAN-gp Transformer has reduced under- and over-provisioning by $2.56\%$ and $1.92\%$, respectively. 
The results clearly indicate that the accurate workload prediction from WGAN-gp Transformer can improve the auto-scaling performance running on real-world cloud platforms.
    
\begin{table}[t]\small
\centering
\resizebox{1\columnwidth}!{
    \begin{tabular}{l|l|c|c}
    \hline
    \bf Workload & \bf Model & \bf $\downarrow$ rate (\%) & \bf $\uparrow$ rate (\%) \\ \hline \hline
    \multirow{2}{*}{Facebook} & LoadDynamics & 40.22 & 10.33 \\ \cline{2-4} 
     & WGAN-gp Tr. & 12.27 & 16.13 \\ \hline
    \multirow{2}{*}{\begin{tabular}[c]{@{}c@{}}{Azure-VM}\\ {-2019}\end{tabular}} & LoadDynamics & 9.63 & 8.60 \\ \cline{2-4} 
     & WGAN-gp Tr. & 7.07 & 6.68 \\ \hline
    \end{tabular}
}
\caption{Auto-scaling evaluation results with VM under-provisioning ($\downarrow$) rates and over-provisioning ($\uparrow$) rates}
\label{table:resource-provisioning-facebook}
\vspace{-1.0em}
\end{table}

\section{Conclusion}
Accurately forecasting user request rates ({\em e.g.,} job arrival rates) benefits optimizing cloud operating costs and guaranteeing application performance goals through predictive VM auto-scaling.
To addresses the problem of job arrival prediction for highly dynamic cloud workloads, we propose a novel time-series forecasting method, called WGAN-gp Transformer, inspired by Transformer network and Wasserstein Generative Adversarial networks. 
When training with an improved WGAN algorithm, the Transformer network accurately captures dynamic patterns in cloud workloads. 
We evaluate WGAN-gp Transformer with various real-world cloud workload datasets against the state-of-the-art LSTM model. 
The results show that our proposed model increases prediction accuracy by up to $5.1\%$, indicating that the attention mechanism in Transformer network can correctly capture relevant information from past workload sequences to make more accurate predictions. 
Furthermore, WGAN-gp Transformer significantly reduces inference time by $5\times$ over the state-of-the-art LSTM-based model. 
Finally, evaluations with real-world cloud deployment show that WGAN-gp Transformer can significantly reduce the under- and over-provisioning rate compared to the baseline model.

\section{Acknowledgments}
Jaewoo Lee's work was supported by the National Science Foundation under Grant No. CNS-1943046.

\small
\bibliography{aaai22.bib}

\begin{thebibliography}{35}
\providecommand{\natexlab}[1]{#1}

\bibitem[{Arjovsky, Chintala, and Bottou(2017)}]{arjovsky2017wasserstein}
Arjovsky, M.; Chintala, S.; and Bottou, L. 2017.
\newblock {Wasserstein Generative Adversarial Networks}.
\newblock In \emph{International Conference on Machine Learning}.

\bibitem[{Calheiros et~al.(2015)Calheiros, Masoumi, Ranjan, and
  Buyya}]{ARIMAWorkPred-TCLOUD15}
Calheiros, R.~N.; Masoumi, E.; Ranjan, R.; and Buyya, R. 2015.
\newblock {Workload Prediction Using ARIMA Model and Its Impact on Cloud
  Applications' QoS}.
\newblock \emph{IEEE Trans. on Cloud Computing}, 3(4).

\bibitem[{Chen et~al.(2011)Chen, Ganapathi, Griffith, and Katz}]{chen2011case}
Chen, Y.; Ganapathi, A.; Griffith, R.; and Katz, R. 2011.
\newblock {The Case for Evaluating MapReduce Performance Using Workload
  Suites}.
\newblock In \emph{IEEE International Symposium on Modelling, Analysis, and
  Simulation of Computer and Telecommunication Systems}.

\bibitem[{Cortez et~al.(2017)Cortez, Bonde, Muzio, Russinovich, Fontoura, and
  Bianchini}]{AzureResourceCentral-SOSP17}
Cortez, E.; Bonde, A.; Muzio, A.; Russinovich, M.; Fontoura, M.; and Bianchini,
  R. 2017.
\newblock {Resource Central: Understanding and PredictingWorkloads for Improved
  Resource Management in Large Cloud Platforms}.
\newblock In \emph{ACM Symp. on Operating Systems Principles}.

\bibitem[{Defazio and Jelassi(2021)}]{defazio2021adaptivity}
Defazio, A.; and Jelassi, S. 2021.
\newblock Adaptivity without Compromise: A Momentumized, Adaptive, Dual
  Averaged Gradient Method for Stochastic Optimization.
\newblock \emph{arXiv preprint arXiv:2101.11075}.

\bibitem[{Ferdman et~al.(2012)Ferdman, Adileh, Kocberber, Volos, Alisafaee,
  Jevdjic, Kaynak, Popescu, Ailamaki, and Falsafi}]{ferdman2012clearing}
Ferdman, M.; Adileh, A.; Kocberber, O.; Volos, S.; Alisafaee, M.; Jevdjic, D.;
  Kaynak, C.; Popescu, A.~D.; Ailamaki, A.; and Falsafi, B. 2012.
\newblock {Clearing the Clouds: A Study of Emerging Scale-out Workloads on
  Modern Hardware}.
\newblock \emph{ACM Sigplan Notices}, 47(4).

\bibitem[{Goodfellow et~al.(2014)Goodfellow, Pouget-Abadie, Mirza, Xu,
  Warde-Farley, Ozair, Courville, and Bengio}]{goodfellow2014generative}
Goodfellow, I.~J.; Pouget-Abadie, J.; Mirza, M.; Xu, B.; Warde-Farley, D.;
  Ozair, S.; Courville, A.; and Bengio, Y. 2014.
\newblock {Generative Adversarial Networks}.
\newblock \emph{arXiv preprint arXiv:1406.2661}.

\bibitem[{Gulrajani et~al.(2017)Gulrajani, Ahmed, Arjovsky, Dumoulin, and
  Courville}]{gulrajani2017improved}
Gulrajani, I.; Ahmed, F.; Arjovsky, M.; Dumoulin, V.; and Courville, A. 2017.
\newblock {Improved Training of Wasserstein GANs}.
\newblock \emph{arXiv preprint arXiv:1704.00028}.

\bibitem[{Hao et~al.(2021{\natexlab{a}})Hao, Jiang, Wang, and
  Kim}]{VMStartupDelay:CLOUD21}
Hao, J.; Jiang, T.; Wang, W.; and Kim, I.~K. 2021{\natexlab{a}}.
\newblock {An Empirical Analysis of VM Startup Times in Public IaaS Clouds}.
\newblock In \emph{{IEEE} International Conference on Cloud Computing}.

\bibitem[{Hao et~al.(2021{\natexlab{b}})Hao, Jiang, Wang, and
  Kim}]{VMStartupDelay:arXiv21}
Hao, J.; Jiang, T.; Wang, W.; and Kim, I.~K. 2021{\natexlab{b}}.
\newblock {An Empirical Analysis of {VM} Startup Times in Public IaaS Clouds:
  An Extended Report}.
\newblock \emph{CoRR}, abs/2107.03467.

\bibitem[{Hochreiter and Schmidhuber(1997)}]{1997-Hochreiter-LSTM}
Hochreiter, S.; and Schmidhuber, J. 1997.
\newblock {Long Short-Term Memory}.
\newblock \emph{Neural Computation}, 9(8): 1735--1780.

\bibitem[{Islam, Venugopal, and Liu(2015)}]{WLBurstiness:SoCC15}
Islam, S.; Venugopal, S.; and Liu, A. 2015.
\newblock {Evaluating the Impact of Fine-scale Burstiness on Cloud Elasticity}.
\newblock In \emph{{ACM} Symposium on Cloud Computing}.

\bibitem[{Jayakumar et~al.(2020)Jayakumar, Lee, Kim, and
  Wang}]{jayakumar2020self}
Jayakumar, V.~K.; Lee, J.; Kim, I.~K.; and Wang, W. 2020.
\newblock A Self-Optimized Generic Workload Prediction Framework for Cloud
  Computing.
\newblock In \emph{IEEE International Parallel and Distributed Processing
  Symposium}.

\bibitem[{Jiang, Chang, and Wang(2021)}]{jiang2021transgan}
Jiang, Y.; Chang, S.; and Wang, Z. 2021.
\newblock {TransGAN: Two Pure Transformers Can Make One Strong GAN, and That
  Can Scale Up}.
\newblock \emph{arXiv preprint arXiv:2102.07074}.

\bibitem[{Jiang et~al.(2011)Jiang, Perng, Li, and Chang}]{ASAP:cloudcom2014}
Jiang, Y.; Perng, C.-S.; Li, T.; and Chang, R.~N. 2011.
\newblock {ASAP: A Self-Adaptive Prediction System for Instant Cloud Resource
  Demand Provisioning}.
\newblock In \emph{IEEE International Conf. on Data Mining}.

\bibitem[{Kim et~al.(2016)Kim, Wang, Qi, and Humphrey}]{EmpEvalWLPred:CLOUD16}
Kim, I.~K.; Wang, W.; Qi, Y.; and Humphrey, M. 2016.
\newblock {Empirical Evaluation of Workload Forecasting Techniques for
  Predictive Cloud Resource Scaling}.
\newblock In \emph{{IEEE} International Conference on Cloud Computing}.

\bibitem[{Kim et~al.(2018)Kim, Wang, Qi, and Humphrey}]{kim2018cloudinsight}
Kim, I.~K.; Wang, W.; Qi, Y.; and Humphrey, M. 2018.
\newblock Cloudinsight: Utilizing a council of experts to predict future cloud
  application workloads.
\newblock In \emph{IEEE International Conference on Cloud Computing}.

\bibitem[{Kim et~al.(2020)Kim, Wang, Qi, and Humphrey}]{CloudInsight:TCC20}
Kim, I.~K.; Wang, W.; Qi, Y.; and Humphrey, M. 2020.
\newblock Forecasting Cloud Application Workloads with CloudInsight for
  Predictive Resource Management.
\newblock \emph{IEEE Transactions on Cloud Computing}, 1--1.

\bibitem[{Kumar et~al.(2018)Kumar, Muthiyan, Gupta, A.D., and
  Nigam}]{2019-Kumar-IJCNN}
Kumar, S.; Muthiyan, N.; Gupta, S.; A.D., D.; and Nigam, A. 2018.
\newblock {Association Learning based Hybrid Model for Cloud Workload
  Prediction}.
\newblock In \emph{International Joint Conference on Neural Networks}.

\bibitem[{Lim and Zohren(2021)}]{lim2021time}
Lim, B.; and Zohren, S. 2021.
\newblock {Time-series Forecasting with Deep Learning: A Survey}.
\newblock \emph{Philosophical Transactions of the Royal Society A}, 379(2194):
  20200209.

\bibitem[{Lin et~al.(2015)Lin, Qi, Yang, and Midkiff}]{lin2015workload}
Lin, H.; Qi, X.; Yang, S.; and Midkiff, S. 2015.
\newblock {Workload-driven VM Consolidation in Cloud Data Centers}.
\newblock In \emph{IEEE International Parallel and Distributed Processing
  Symposium}.

\bibitem[{Loff and Garcia(2014)}]{Vadara:cloudcom2014}
Loff, J.; and Garcia, J. 2014.
\newblock {Vadara: Predictive Elasticity for Cloud Applications}.
\newblock In \emph{IEEE International Conference on Cloud Computing Technology
  and Science}.

\bibitem[{Mao and Humphrey(2011)}]{MingAutoscaling:SC11}
Mao, M.; and Humphrey, M. 2011.
\newblock {Auto-scaling to Minimize Cost and Meet Application Deadlines in
  Cloud Workflows}.
\newblock In \emph{International Conference on High Performance Computing
  Networking, Storage and Analysis}.

\bibitem[{Mao and Humphrey(2012)}]{VMStartupDelay:CLOUD12}
Mao, M.; and Humphrey, M. 2012.
\newblock {A Performance Study on the {VM} Startup Time in the Cloud}.
\newblock In \emph{{IEEE} International Conference on Cloud Computing}.

\bibitem[{Nguyen, Klein, and Elmroth(2019)}]{2019-Nguyen-CCGrid}
Nguyen, C.; Klein, C.; and Elmroth, E. 2019.
\newblock {Multivariate LSTM-Based Location-Aware Workload Prediction for Edge
  Data Centers}.
\newblock In \emph{Int'l Symposium on Cluster, Cloud and Grid Computing}.

\bibitem[{Roy, Dubey, and Gokhale(2011)}]{roy2011efficient}
Roy, N.; Dubey, A.; and Gokhale, A. 2011.
\newblock {Efficient Autoscaling in the Cloud Using Predictive Models for
  Workload Forecasting}.
\newblock In \emph{IEEE International Conference on Cloud Computing}.

\bibitem[{Rzadca et~al.(2020)Rzadca, Findeisen, Swiderski, Zych, Broniek,
  Kusmierek, Nowak, Strack, Witusowski, Hand, and
  Wilkes}]{Autoscale:Eurosys2020}
Rzadca, K.; Findeisen, P.; Swiderski, J.; Zych, P.; Broniek, P.; Kusmierek, J.;
  Nowak, P.; Strack, B.; Witusowski, P.; Hand, S.; and Wilkes, J. 2020.
\newblock {Autopilot: workload autoscaling at Google}.
\newblock In \emph{ACM European Conference on Computer Systems}.

\bibitem[{Shen et~al.(2011)Shen, Subbiah, Gu, and Wilkes}]{shen2011cloudscale}
Shen, Z.; Subbiah, S.; Gu, X.; and Wilkes, J. 2011.
\newblock {Cloudscale: Elastic Resource Scaling for Multi-tenant Cloud
  Systems}.
\newblock In \emph{ACM Symposium on Cloud Computing}.

\bibitem[{Tsai et~al.(2019)Tsai, Bai, Yamada, Morency, and
  Salakhutdinov}]{tsai2019transformer}
Tsai, Y.-H.~H.; Bai, S.; Yamada, M.; Morency, L.-P.; and Salakhutdinov, R.
  2019.
\newblock Transformer Dissection: A Unified Understanding of Transformer's
  Attention via the Lens of Kernel.
\newblock \emph{arXiv preprint arXiv:1908.11775}.

\bibitem[{Vaswani et~al.(2017)Vaswani, Shazeer, Parmar, Uszkoreit, Jones,
  Gomez, Kaiser, and Polosukhin}]{vaswani2017attention}
Vaswani, A.; Shazeer, N.; Parmar, N.; Uszkoreit, J.; Jones, L.; Gomez, A.~N.;
  Kaiser, L.; and Polosukhin, I. 2017.
\newblock {Attention is All You Need}.
\newblock \emph{arXiv preprint arXiv:1706.03762}.

\bibitem[{Wilkes(2011)}]{clusterdata:Wilkes2011}
Wilkes, J. 2011.
\newblock More {Google} cluster data.
\newblock
  \url{http://googleresearch.blogspot.com/2011/11/more-google-cluster-data.html}.
\newblock Accessed: 2021-09-01.

\bibitem[{Wood et~al.(2008)Wood, Cherkasova, Ozonat, and
  Shenoy}]{wood2008profiling}
Wood, T.; Cherkasova, L.; Ozonat, K.; and Shenoy, P. 2008.
\newblock {Profiling and Modeling Resource Usage of Virtualized Applications}.
\newblock In \emph{ACM/IFIP/USENIX International Middleware Conference}.

\bibitem[{Wu et~al.(2020)Wu, Xiao, Ding, Zhao, Wei, and
  Huang}]{wu2020adversarial}
Wu, S.; Xiao, X.; Ding, Q.; Zhao, P.; Wei, Y.; and Huang, J. 2020.
\newblock {Adversarial Sparse Transformer for Time Series Forecasting}.
\newblock In \emph{Advances in Neural Information Processing Systems}.

\bibitem[{Xu et~al.(2015)Xu, Wang, Chen, and Li}]{xu2015empirical}
Xu, B.; Wang, N.; Chen, T.; and Li, M. 2015.
\newblock {Empirical Evaluation of Rectified Activations in Convolutional
  network}.
\newblock \emph{arXiv preprint arXiv:1505.00853}.

\bibitem[{Yadwadkar, Ananthanarayanan, and Katz(2014)}]{Wrangler-SoCC14}
Yadwadkar, N.~J.; Ananthanarayanan, G.; and Katz, R. 2014.
\newblock {Wrangler: Predictable and Faster Jobs using Fewer Resources}.
\newblock In \emph{{ACM} Symposium on Cloud Computing}.

\end{thebibliography}

\end{document}